\documentclass[10pt]{article}

\usepackage{times}
\usepackage{graphicx} 
\graphicspath{{./pictures/}}
\DeclareGraphicsExtensions{.pdf,.jpeg,.png}

\usepackage{subfigure} 

\usepackage{algpseudocode}
\usepackage{algorithm}
\usepackage{algorithmicx}

\usepackage[table]{xcolor}
\usepackage{booktabs}
\usepackage{multirow}
\usepackage{multicol}

\usepackage[cmex10]{amsmath} 
\usepackage{amsthm}
\usepackage{amssymb}

\usepackage{url}

\newcommand\tstrut{\rule{0pt}{2.4ex}}
\newcommand\bstrut{\rule[-1.0ex]{0pt}{0pt}}

\newcolumntype{a}{>{\columncolor{gray!35}}c}

\title{Real-world Object Recognition with Off-the-shelf Deep Conv Nets: \\How Many Objects can iCub Learn?}

\author{Giulia Pasquale  \thanks{
       iCub Facility, Istituto Italiano di Tecnologia,
       Via Morego, 30,
       16100, Genova, Italy, ({\tt giulia.pasquale@iit.it,lorenzo.natale@iit.it})} \ 
 \and
Carlo Ciliberto \thanks{
       Laboratory for Computational and Statistical Learning, Istituto Italiano di Tecnologia,
       Via Morego, 30,
       16100, Genova, Italy, ({\tt cciliber@mit.edu,lrosasco@iit.it})} \ 
 \and
Francesca Odone  \thanks{
       DIBRIS, Universit\`a di Genova,
       Via Dodecaneso, 35,
       16146, Genova, Italy, ({\tt francesca.odone@unige.it})} \ 
 \and
Lorenzo Rosasco $^\dagger$ $^\ddagger$  \
 \and
Lorenzo Natale $^*$
}
\date{}

\begin{document}

\maketitle
\thispagestyle{empty}
\pagestyle{empty}

\begin{abstract}

The ability to visually recognize objects is a fundamental skill for robotics systems. Indeed, a large variety of tasks involving manipulation, navigation or interaction with other agents, deeply depends on the accurate understanding of the visual scene. Yet, at the time being, robots are lacking good visual perceptual systems, which often become the main bottleneck preventing the use of autonomous agents for real-world applications.

Lately in computer vision, systems that learn suitable visual representations and based on multi-layer deep convolutional networks are showing remarkable performance in tasks such as large-scale visual recognition and image retrieval. To this regard, it is natural to ask whether such remarkable performance would generalize also to the robotic setting.

In this paper we investigate such possibility, while taking further steps in developing a computational vision system to be embedded on a robotic platform, the iCub humanoid robot. In particular, we release a new dataset ({\sc iCubWorld28}) that we use as a benchmark to address the question: {\it how many objects can iCub recognize?} Our study is developed in a learning framework which reflects the typical visual experience of a humanoid robot like the iCub. Experiments shed interesting insights on the strength and  weaknesses of current computer vision approaches applied in real robotic settings.
\end{abstract}

\section{Introduction}\label{sec:intro}

Visual perception is arguably one of the most important sensory channels for robotic systems that should operate in human environments. Indeed, the lack of good visual information becomes a major bottle neck for almost any task in which the robotic agent might be engaged, from simple manipulation to complex behaviors implying planning and navigation. 

In recent years, computational vision systems have witnessed tremendous progress, especially in the context of object recognition. Such a progress has been mainly driven by the development of machine learning methods for representing and classifying images, based on multi-layers (deep) architectures (see \cite{yang09,csurka04,mutch08} and, more recently, \cite{sermanet14,Krizhevsky12,jia2014caffe}). An important reason for the rapid evolution of this kind of systems was the acquisition of large public data-sets on which to train and benchmark the performance of new solutions, e.g. Caltech$256$~\cite{caltech256}, PascalVOC~\cite{everingham10} and ImageNet LSVRC~\cite{ILSVRCarxiv14}. All these datasets are essentially  tailored to image retrieval problems and indeed  this is the kind of task on  which the performance of many vision systems have been ultimately tested. It is then natural to ask to which extent these new developments can impact robotics systems where the vision tasks of interest are different from the typical retrieval scenario.

\begin{figure}[t]
\begin{center}
   \includegraphics[width=0.65\columnwidth]{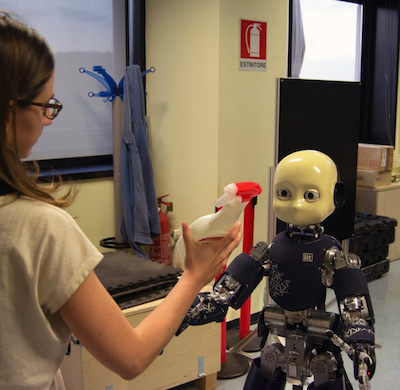}
\end{center}
   \caption{Setup used to collect the {\sc iCubWorld28} dataset.}
\label{fig:setup}
\end{figure}

The iCub humanoid~\cite{icub} (see Figure~\ref{fig:setup}) offers an ideal platform to address the above question that we began to  investigate in~\cite{ciliberto13,fanello13}. In particular, we started collecting and making available a dataset ({\sc iCubWorld}\footnote{ \texttt{ \url{http://www.iit.it/en/projects/data-sets.html}}}) that reflects the typical visual experience of iCub and testing different solutions for visual recognition.  Our preliminary results confirmed on the one hand the potential of recently proposed systems, and on the other highlighted the challenges posed by the specific robotics context -- in particular the lack of accurate supervision. 

The current paper builds on our previous work to take a further step in the development of a computational vision system for the iCub. In particular, in this paper we conduct an empirical study aimed at answering the question: 
\begin{center}
{\it ``How many objects can iCub recognize today?''}
\end{center}
We consider this problem within the Human-Robot Interaction scenario proposed in~\cite{ciliberto13} for the acquisition of the {\sc iCubWorld} dataset. In the current work, a human teacher shows $28$ different objects to the iCub, verbally annotating them using a speech recognition system to provide labeling. The same procedure is repeated for four consecutive days, leading to the acquisition of a new dataset, dubbed {\sc iCubWorld28}.

We conjugate the above broad question in further sub-problems that we describe in Sec.~\ref{sec:questions} and address empirically in Sec.~\ref{sec:answers}. We devote Sec.~\ref{sec:methods} to provide some background on the general problem of learning visual representations. In Sec.~\ref{sec:setup} we first review the robotic application that we designed to perform the acquisition of {\sc iCubWorld28} and then we describe the image representation pipeline employed in our experiments. Finally Sec.~\ref{sec:conclusion} concludes the work, laying the foundations for future research.

\section{An ideal robotic visual recognition system}\label{sec:questions}

By asking ``{\it How many objects can iCub recognize?}'', in this work we aim to investigate the problem of achieving human-like visual recognition capabilities in robotics. To address this question we divide our analysis into multiple points:\\

\noindent {\bf \em Reliability}
In order to be reproducible, our analysis will be performed off-line on a visual recognition dataset directly acquired from the robot cameras, {\sc iCubWorld28}. However, in order to generalize the recognition performance observed on such a benchmark, we will need a measure able to quantify the confidence with which we can expect such results to hold also in the real-world application. In Sec.~\ref{sec:increasing_classes} we propose a possible approach to this problem.\\

\noindent {\bf \em Contextual Information.}
The robotic setting offers a great deal of contextual information that could be incorporated in the learning system to improve recognition performance. For instance, by observing an object from different points of view, the robot could be able to better disambiguate between different classes. Typically, contextual information is not available in standard computer vision settings and therefore is unclear in general how to employ it in recognition. In Sec.~\ref{sec:time} we start addressing this question.\\

\noindent {\bf \em Learning incrementally.}
A human-like artificial system should be able to learn a richer model of the world as new observations become available. Specifically, it is natural to expect that the visual recognition system of a humanoid robot should benefit from the incorporation of visual data acquired on multiple occasions, such as training sessions across multiple days. A preliminary analysis of such an incremental setting is performed in Sec.~\ref{sec:incremental} and represents a first step towards a true life-long learning system that continuously updates its internal model of the world.\\

\noindent {\bf \em Self-Supervision.} 
Ideally, the interaction between a human and a robot should take place along natural communication channels (for the human), such as speech or vision. Clearly, such a scenario limits the amount of supervision that a human teacher can provide to the robot. For instance, in the human-robot application considered in this work, images cannot be manually segmented around the object of interest and therefore the system has to rely on so-called ``weak'' or ``self-'' supervised strategies, such as motion segmentation, to eliminate, at least partially, the visual distractors (e.g. background or other objects). In Sec.~\ref{sec:background} we investigate this problem, evaluating the impact of having a finer (or coarser) segmentation for the images in {\sc iCubWorld28}.

\section{Learning to Represent and Classify Objects}\label{sec:methods}

Modern vision algorithms for recognition/categorization rely on machine learning routines to identify a suitable {\it representation} for the visual data. Ideally, such a representation (usually encoded in a real vector of finite dimension) should be on one hand {\it discriminative}, in the sense that images depicting different objects should be easily separable, while on the other hand being {\it invariant} to physical transformations of the scene (such as translations, rotations or deformations) that do not affect the actual object class.

These methods are typically composed of two or more layers alternating convolution and non-linear mappings of local image patches; the training process usually consists in the optimization of the weights in the convolution stage with respect to a given loss function (e.g. reconstruction error) separately or jointly for all layers. Several approaches where proposed to learn the local filters at the convolution level, such as Bag of Words~\cite{csurka04}, Sparse Coding~\cite{yang09}, Fisher Vectors~\cite{Perronnin10} or HMAX~\cite{serre07}. Once the representation map has been learned, each novel image is mapped to the the new space where a classifier is trained using standard techniques for supervised learning such as SVM~\cite{scholkopf02}, Regularized Least Squares (RLS)~\cite{hastie09} or Neural Networks (NN)~\cite{bishop06}. 

Lately, the availability of extremely large image datasets and parallel computing resources, such as general-purpose GPUs, has made possible to train deep architectures as Convolutional Neural Networks (CNNs) on all layers simultaneously. According to recent empirical evidence~\cite{donahue2013decaf,zeiler13visualizingcnn,Chatfield14,razavian14cnnfeat} it appears that architectures trained on such a rich amount of visual information are able to develop extremely powerful representations, and therefore can be used also on novel datasets as a ``black-box'' for image description. This approach is particularly appealing and is the one evaluated in this paper. Indeed, at the time being effectively training such complex architectures from scratch still requires very large numbers of examples, high computational times and, not least, the know-how to accurately tune their parameters, all factors that are not trivial in robotics settings where the application context can be not known a-priori.

Other lines of research for visual recognition are based on keypoints matching techniques~\cite{lowe04,philbin2008lost,collet2011structure,crowley2014state}. Although often employed in robotics settings~\cite{Collet2011,collet2009,muja2011rein}, these approaches are not particularly suited to applications where supervision is not accurate. Indeed, we previously observed in~\cite{ciliberto13} that, when employed in natural Human-Robot Interaction scenarios where supervision is weak, the performance of these methods degrades remarkably. Hence in this work we do not cover these and related approaches and instead we focus on learning representation methods.

\begin{figure}[t]
\begin{center}
   \includegraphics[width=\columnwidth]{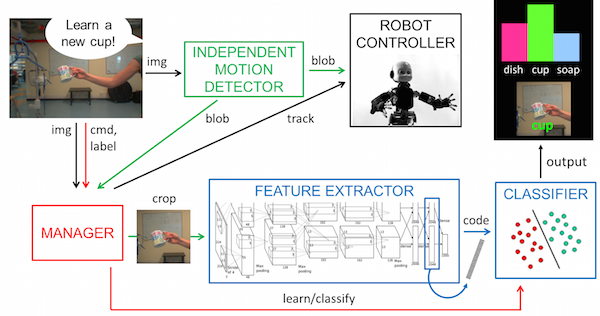}
\end{center}
   \caption{The visual recognition system adopted in this work and currently implemented on iCub.}

\label{fig:pipeline}
\end{figure}

\section{Setup and Acquisition}\label{sec:setup}

In this section we describe the image acquisition protocol used to collect the {\sc iCubWorld28} dataset, as well as the implementation details of the visual recognition framework employed for the experimental analysis discussed in Sec.~\ref{sec:answers}. \\

\noindent{\bf Setup.} The application setup we employed in this work is analogous to the one described in~\cite{ciliberto13} and we briefly outlined it in the introduction of this paper: a human supervisor is standing in front of the iCub robot and shows it different objects while verbally providing the class annotation (Figure~\ref{fig:setup} depicts a typical acquisition setting). Exploiting independent motion detection routines~\cite{ciliberto12}, the robot tracks the novel object while acquiring images at $33$hz. The independent motion detection algorithm allows to perform an approximate localization of the object, effectively reducing the image size from $320 \times 240$ pixels to a mean of $\sim 120 \times 120$ pixels (See Fig.~\ref{fig:pipeline} for an example). Cropped images are then processed by a representation module (see Sec.~\ref{sec:methods}) that encodes the visual information into a single vector or descriptor that will then be used for classification (Figure~\ref{fig:pipeline}).\\

\begin{figure}[t]
\begin{center}
   \includegraphics[width=\columnwidth]{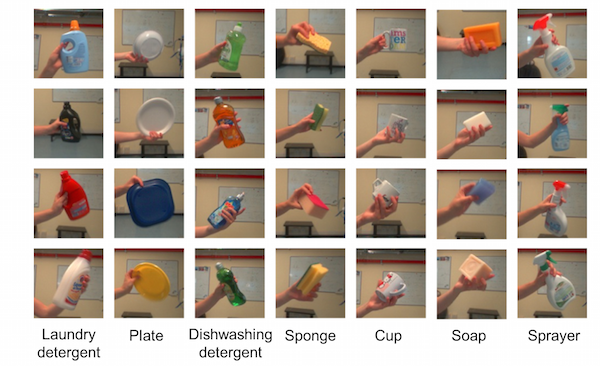}
\end{center}
   \caption{Example images from one of the $4$ datasets comprising {\sc iCubWorld28}. As can be seen in the Figure, each dataset is composed by $28$ objects organized into $7$ categories.}
\label{fig:dataset}
\end{figure}

\noindent{\bf Acquisition.} Within the setting described above, we collected the {\it {\sc iCubWorld28}} dataset which comprises images of $28$ distinct objects evenly organized into $7$ categories (see Figure~\ref{fig:dataset}). For each object in the dataset, we acquired a separate train and test sets during sessions of $20$ seconds each. We reduced the acquisition frequency by a factor of $3$ (i.e. acquiring one image around every $0.09$ seconds) to lower the computational costs of the learning process. Thus, after each session, we collected $220$ train and $220$ test images for each of the $28$ objects. To assess the incremental learning performance of the iCub visual recognition system (see the discussion in Sec.~\ref{sec:incremental}) we repeated this same acquisition protocol for $4$ consecutive days, ending up with four datasets (Day $1$, to $4$) of more than $12$k images each and $~50$k images in total. We will make this release available for the community at the same web address of the previous {\sc iCubWorld}.\\

\noindent{\bf Extracting visual representation.} 
To extract visual representations of images acquired from iCub's cameras, in this work we employed a CNN originally trained on the ImageNet dataset~\cite{ILSVRCarxiv14}. Specifically we employed a model provided in Caffe's library~\cite{jia2014caffe}, \textit{BVLC Reference CaffeNet}, which is available on-line and is based on the well-established network proposed in~\cite{Krizhevsky12}. Following the strategy proposed in~\cite{donahue2013decaf, razavian14cnnfeat, Chatfield14}, we employed the CNN as a black-box module that takes images in input and returns their corresponding vector representations in output.\\



\noindent{\bf Learning.} 
In visual recognition settings, the typical approach to classification is to employ so-called {\it supervised learning} methods such as Support Vector Machines or Regularized Least Squares (RSL). In this work we rely on the GURLS~\cite{tacchetti13} machine learning library to perform RLS. Indeed, as empirically observed from previous work on the iCub~\cite{ciliberto13}, RLS exhibited comparable or even better results than Support Vector Machines (using the liblinear~\cite{liblin} library). Moreover, the rank-one update rule for matrix inversion~\cite{golub12} provides a natural variant of the classic RLS algorithm to the setting in which training data is provided incrementally to the system (also the incremental RLS algorithm is implemented in the GURLS library). Clearly, this is a typical scenario in robotics applications and, as already mentioned in Sec.~\ref{sec:questions} is a topic of interest in this work (see Sec.~\ref{sec:incremental}).


\section{A data sheet of iCub's visual recognition capabilities}\label{sec:answers}

In this section we empirically address the questions raised in Sec.~\ref{sec:questions}, with the aim of providing the reader with an ideal ``data sheet'' of iCub's current visual capabilities. The guiding principle of our analysis is to answer the generic (and intentionally fuzzy) question ``{\it How many object can iCub recognize?}'' where, with the word ``recognize'', we refer to human-level visual capabilities. Indeed, in realistic robotic applications we need reliable perceptual systems that, at least for limited sets of objects, are virtually infallible.

In order to investigate this problem, in Sec.~\ref{sec:increasing_classes} we first introduce and discuss a possible way to measure the confidence with which the classification accuracy achieved by systems trained on our benchmark dataset, {\sc iCubWorld28}, is expected to generalize during a generic run of the human-robot interaction application described in Sec.~\ref{sec:setup}. Then, in the following Sec.~\ref{sec:time},~\ref{sec:incremental} and~\ref{sec:background} we consider natural approaches to improve recognition in the robotic settings, identifying possible future directions for research. In Sec.~\ref{sec:comparison_previous_work} we briefly report a comparison of different visual recognition systems for reference, while in Sec.~\ref{sec:final_answer} we provide a preliminary answer to the question motivating this work.

\subsection{Reliability and Scalability}\label{sec:increasing_classes}

Ideally, a reliable recognition system should be robust with respect to set of objects it has to discriminate. In other words, we would like the classification performance of a predictor to not vary dramatically when we change the set of classes on which it is trained/tested. This problem is particularly relevant to this work since the main goal of our analysis -- although limited to the dataset of $28$ objects described in Sec.~\ref{sec:setup} -- is to offer insights on the expected recognition capabilities of iCub for any choice of objects. 

\begin{figure}[t]
\begin{center}
   \includegraphics[width=\columnwidth]{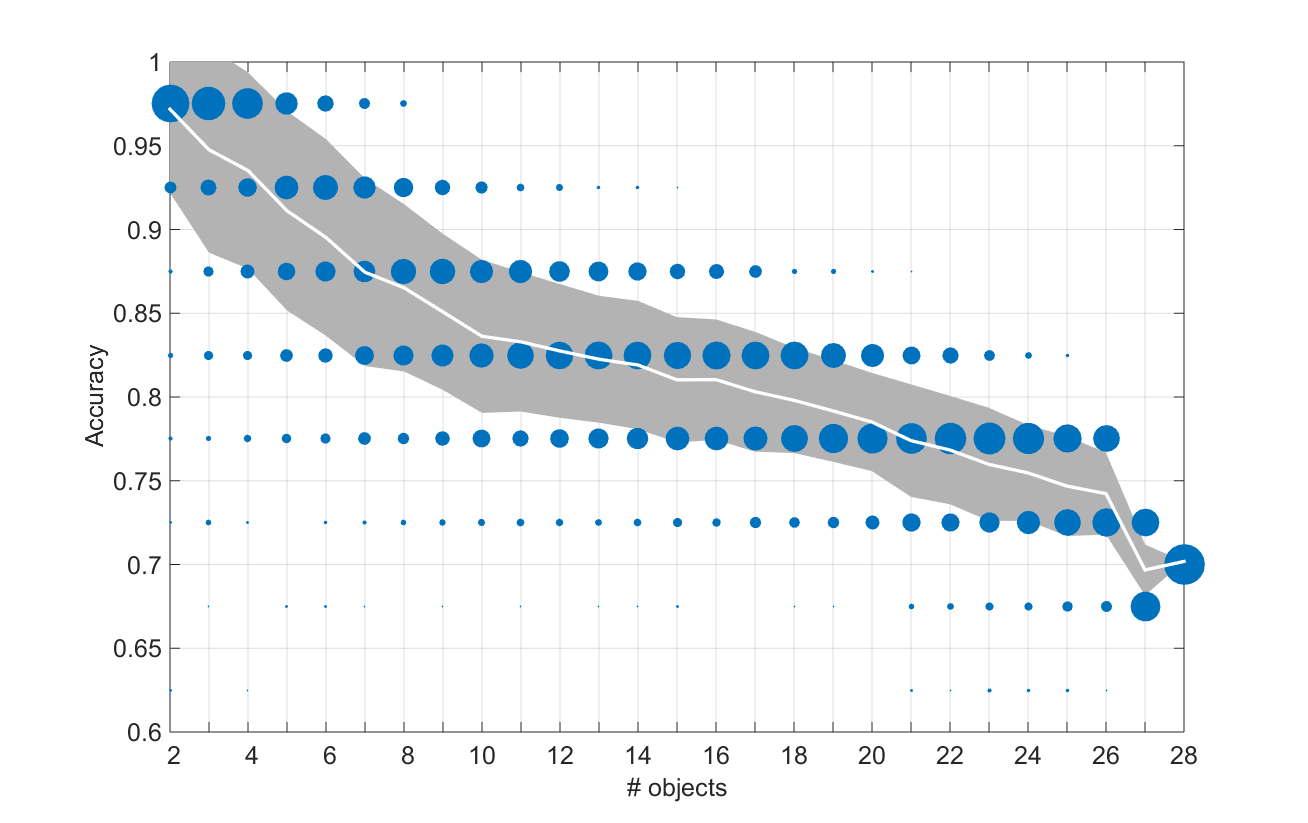}
\end{center}
   \caption{Empirical estimation of the probability distribution $P(acc = A|t)$ for a predictor trained on a random set of $t$ objects to have accuracy $A$.}
\label{fig:acc_vs_t}
\end{figure}

Therefore, to quantitatively measure the reliability of the visual system currently available on iCub, we performed multiple classification experiments for different subsets of classes in {\sc iCubWorld28} for the dataset corresponding to Day $1$. More precisely, for any $t=2,\dots,26$ we randomly selected $\sim400$ different combinations of $t$ object classes among the available $28$ (to avoid the combinatorial explosion of $\tbinom{28}{t}$ experiments) and trained/tested the learning system described in Sec.~\ref{sec:setup} on the corresponding reduced datasets. As a measure of performance for the resulting predictor we computed its {\it average accuracy}, namely the ratio of correct guesses with respect to the cardinality of the whole test set. For a fixed $t$, we interpreted the accuracy measured for each individual experiment as one observation sampled from $P(acc = A | t)$, namely the conditional probability that a predictor trained on a randomly chosen set of $t$ classes would achieve accuracy equal to a value $A$ between $0$ and $1$. In Fig.~\ref{fig:acc_vs_t} we report the empirical estimation of this distribution  together with the associated empirical mean (white curve) and one the standard deviation (gray region). Specifically, each column in the plot approximates $P(acc = A | t)$ as the normalized histogram of the accuracies measured for a fixed $t$ and is depicted as a vertical sequence of balls with radius directly proportional to the corresponding bin value.

Apart from the expected drop in accuracy that we observe when the cardinality of the multi-class problem increases, this analysis provides us with useful insights: first notice that the slope of the mean accuracy reported in Fig.~\ref{fig:acc_vs_t} (white curve) experiences a remarkable decrease as the number of classes increases (e.g. after $t=10$), suggesting that such a negative effect should become less and less disrupting as we learn new objects. To confirm this trend and further investigate the behavior of such a recognition system, in the near future we will extend our analysis to a larger version of {\sc iCubWorld28}, containing more object classes and categories. Indeed, one of the guiding principles behind the {\sc iCubWorld} project is actually to collect a dataset in constant expansion whose incremental growth would retrace the natural experience of a physical agent that explores an unknown environment and discovers new objects.

\begin{figure}[t]
\begin{center}
   \includegraphics[width=\columnwidth]{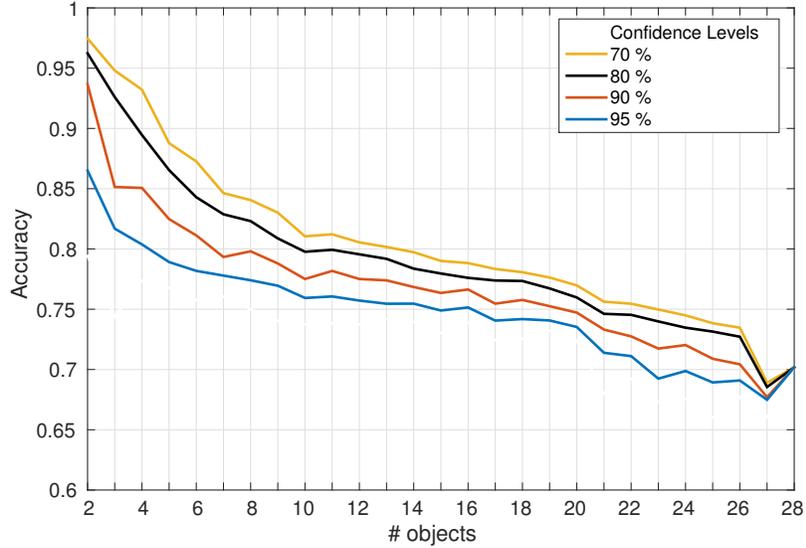}
\end{center}
   \caption{Confidence intervals for predictors trained on a randomly sampled set of objects. For a fixed number of objects $t$, the value on a curve $C$ represents the minimum accuracy that we are guaranteed to achieve with the trained predictor, with confidence $C$.}
\label{fig:confidence}
\end{figure}


Second, notice that for each fixed cardinality $t$, the distribution of accuracies $P(acc = A | t)$, measured across the multiple trials, is clearly concentrated around its mean. More specifically, this means that in general we can expect with high confidence that a predictor trained on a randomly selected set of $t$ objects would have accuracy between $\pm 5\%$ of the mean of $P(acc = A | t)$. This offers a useful perspective on what recognition performance we should expect during a typical run of the human-robot interaction application described in Sec.~\ref{sec:setup}, ideally for a any random selection of $t$ objects. To better quantify the expected capability of the system to generalize its performance, in Fig.~\ref{fig:confidence} we report the minimum accuracy that we are guaranteed to achieve within specified levels of confidence. In this context, the confidence $c(A,t)$ for a given accuracy $A$ and number of classes $t$ was measured as
\begin{equation}\label{eq:confidence}
c(A,t) = \int_A^1 P(acc = a|t) \ \mathrm{d}a
\end{equation}
and in Fig.~\ref{fig:confidence} we report the confidence level curves $c(A,t) = C$ for different values of $C$. Such curves denote the minimum accuracy $A$ guaranteed for a classifier trained on a random set of $t$ objects. To better understand the implications of this analysis, let us consider for instance the Blue curve in Fig.~\ref{fig:confidence}, related to $95\%$ and passing by $t=15$ and $A=0.75$: with high probability ($95\%$) and for a random choice of $15$ objects, the resulting predictor is guaranteed to achieve at least $0.75$ classification accuracy. 

This result, and its corresponding visualization in Fig.~\ref{fig:confidence}, is of particular use from a practical perspective since it can be employed as a reference ``data sheet'' to train the iCub. Indeed, depending on the desired confidence $C$ and the number $t$ of objects we want the robot to discriminate, Fig.~\ref{fig:confidence} informs us what is the approximate level of accuracy that we can expect to achieve with the classifier that we will train.


\begin{figure}[t]
\begin{center}
  \includegraphics[width=\columnwidth]{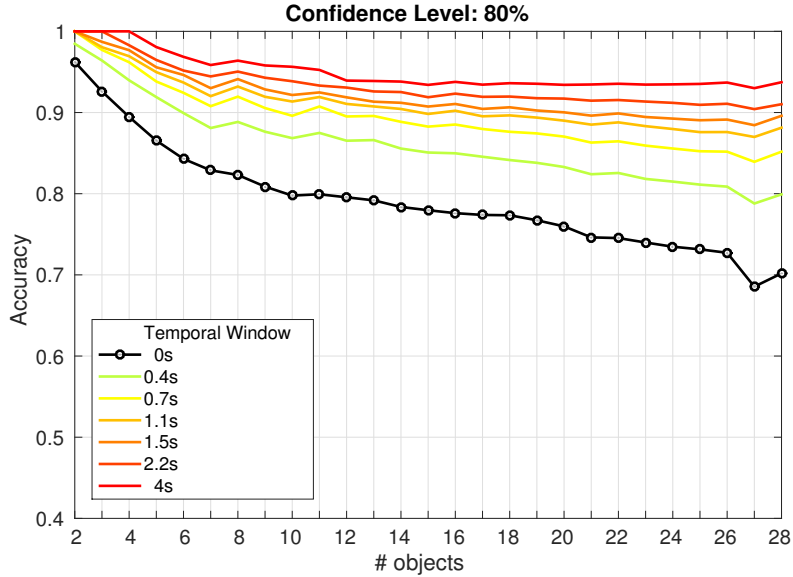}
\end{center}
   \caption{Improvement of the classification accuracy with respect to an increasingly large temporal filtering window. Results are shown for fixed confidence level $C=80\%$ (see Eq.~\eqref{eq:confidence} and Fig.~\ref{fig:confidence}).}
\label{fig:prob_vs_t_withfilter}
\end{figure}

\subsection{Exploiting Contextual Information}
\label{sec:time}

The classification performances reported in Fig.~\ref{fig:confidence} are clearly not comparable to the human-level accuracy that we would expect on the problem considered. Indeed, even for relatively low confidence values such as $80\%$ (Black curve), we observe a fast decay of the guaranteed accuracy, which falls under the $0.9$ threshold just after $4$ objects.

A viable approach to mitigate this problem relies on noticing that the robotic setting offers a great deal of prior and contextual information that could remarkably improve performance. To this regard, let us consider the natural assumption that the class of an object does not change while the robot observes it from multiple points of view. In such a setting, given a set of $w$ images (acquired from different viewpoints around the object of interest) and a trained classifier with a (per-frame) accuracy $A$, we can consider a new classification rule that combines the individual predictions on the set of $w$ frames into a global label. For instance, if we assume that the $w$ images are sampled i.i.d., we have that the rule returning the label that occurred at least $50\%+1$ times would correctly classify the object with probability (or accuracy)
\begin{equation}\label{eq:probability}
P = \sum_{k=\lfloor w/2 \rfloor + 1}^w \binom{w}{k} A^k (1-A)^{w-k}.
\end{equation}
In principle, this strategy could be extremely beneficial: suppose for instance that the trained predictor has a per-frame accuracy of $A=0.7$. Then, even for small sets (or windows) of just $3$ images we would have improved classification accuracy of $0.78$, while for a larger $w=21$ we would achieve an impressive $0.97$.

We evaluated the approach described above on {\sc iCubWorld28}. In particular, since in our setting the samples are acquired as a stream of consecutive images and we are interested in on-line recognition, we chose to classify windows selecting the current frame together with the previous $w-1$ ones. This approach could be interpreted as a sort of label-filtering process that suppresses ``flickering'' one-frame misclassification. Clearly, in this case Eq.~\eqref{eq:probability} represents only an upper bound to the actual improvement that we can expect, since the i.i.d. assumption never holds (consecutive images in a stream are of course always correlated).

Figure~\ref{fig:prob_vs_t_withfilter} reports the effect of the label-filtering approach on the confidence curve associated to $C=80\%$ introduced in Fig.~\ref{fig:confidence}. We varied the size of the temporal windows from $0$ (instantaneous) to $4$ seconds, corresponding to a range of $w$ between $1$ and $50$ frames. Notice that even in this non i.i.d. scenario, the system performance clearly benefits from smoothing, in particular when several classes are considered. Probably this is due to the fact that, as the number of object to discriminate grows, the chance of short-lived ``one-frame'' misclassification increases proportionally. 

\subsection{Incremental Learning: A week (almost) with iCub}\label{sec:incremental}

The temporal filtering strategy considered in Sec.~\ref{sec:time} leads to an impressive boost in recognition accuracy. However, if we consider the original goal of achieving human-level performance on {\sc iCubWorld} (say, for reference, $0.98$ accuracy), we notice from Fig.~\ref{fig:prob_vs_t_withfilter} that even for a relatively low confidence value of $80\%$ the system is still lacking a significant accuracy gap. To this regard, in this section we take into account another aspect of robotics settings that could in principle improve the recognition capabilities of the system, namely the ability to learn incrementally.

\begin{figure}[t]
\begin{center}
  \includegraphics[width=\columnwidth]{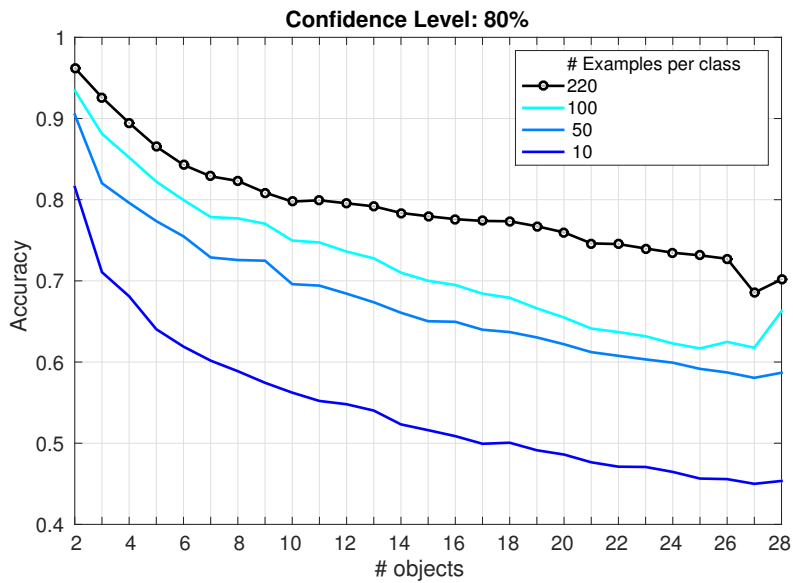}
\end{center}
   \caption{Classification accuracy for fixed level of confidence $C=80\%$ (see Eq.~\eqref{eq:confidence}) and an incremental number of training examples per class.}
\label{fig:incremental_single_day}
\end{figure}

Indeed, the robotic scenario is naturally suited to life-long learning applications. Specifically, in visual recognition settings, novel training evidence could be provided to the robot incrementally (and in principle, indefinitely) in order to update its knowledge as the task requires. A first result, that empirically quantifies the importance of learning incrementally and motivates the experimental analysis of this section, is reported in Fig.~\ref{fig:incremental_single_day}. We consider the experimental setting introduced in Sec.~\ref{sec:increasing_classes} and report the curve associated to $80\%$ confidence for classifiers trained on an incremental number of examples per class. As can be noticed, the incremental growth of the training data has a remarkable impact on the overall classification performance and opens the question of what would be the long-term effects of such a learning process on the system's recognition capabilities.

\begin{figure}[t]
\begin{center}
  \includegraphics[width=\columnwidth]{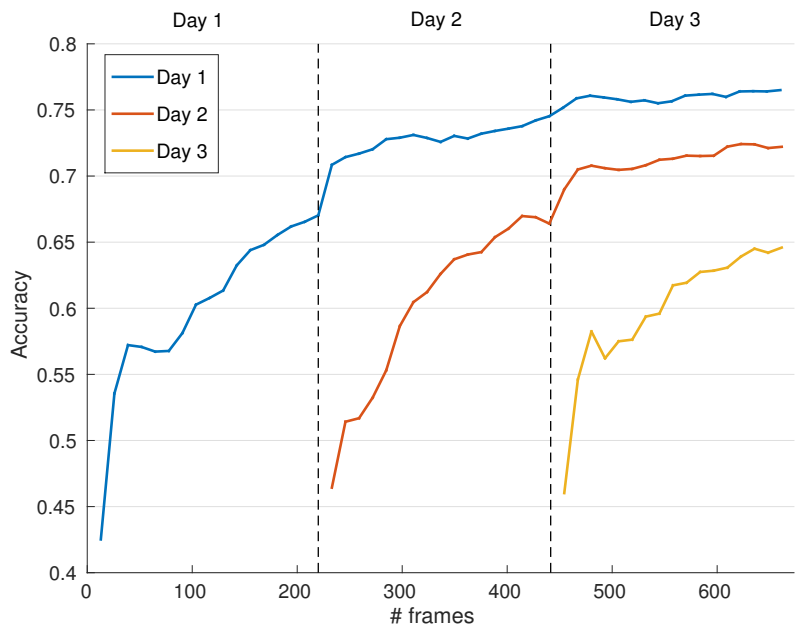}
\end{center}
   \caption{Incremental learning on {\sc iCubWorld28}. Blue, Orange and Yellow curves identify the classification accuracy of predictors trained incrementally starting from, respectively, Day $1$, Day $2$ and Day $3$. We used the test set from Day $4$ to assess the generalization performance of the classifiers.}
\label{fig:acc_vs_nuples}
\end{figure}

We recall that {\sc iCubWorld28} is a dataset collected during $4$ separate days and that for each day both a training and test set were acquired. We further recall that all experiments discussed so far were performed on a single day of {\sc iCubWorld28}, say Day $1$. To the purpose of studying the impact of incremental learning on visual recognition, in the following we will take into account also to the remaining $3$ days. In particular, we considered the learning setting in which we trained a classifier incrementally on the training sets of the first three days of {\sc iCubWorld28} and then evaluated it on the tests set of the fourth ``unseen'' day. To reduce the amount of computations we focused only on the problem of correctly classifying the $28$ objects in the dataset and report the measured accuracy in Fig.~\ref{fig:acc_vs_nuples} for classifiers trained starting respectively from Day $1$ (Blue), Day $2$ (Orange) and Day $3$ (Yellow). On one hand, we notice that when provided only with training data acquired from a single day, the incremental learning accuracy exhibited by predictors follows a remarkably similar pattern for all days, suggesting the the three datasets contain a similar amount of information. On the other hand, we observe that while all these curves seem to saturate around $\sim0.65$ accuracy, adding data from a new day allows to overcome such limitation, improving the overall system performance (here we refer to the ``jumps'' observed for both the Blue and Orange curves as they switch between days).

The results reported in Fig.~\ref{fig:acc_vs_nuples} seem to suggest that training across multiple days is more beneficial than training during a single session because it exposes the system to less redundant information. To confirm this observation we considered a further experimental scenario where we compared the performance of a predictor trained on data acquired from all days with the accuracy achieved by other four classifiers, each trained on a different day of {\sc iCubWorld28} taking the first $100$ examples per class. In order to compare problems of identical dimension, the ``mixed'' dataset was created by taking the first $25$ samples (per class) from the training set of each day. Table~\ref{tab:redundancy} reports the resulting classification accuracy tested separately on each day. In line with the original intuition, we notice that predictor trained on the mixed dataset clearly outperforms the others on average. However, it is of particular interest to observe that even on a single day basis, the predictor trained on all days (and thus less exposed to redundant information) outperforms predictors trained and tested on the same day.


\begin{table}[t]
\begin{center}

\begin{tabular}{rrcccca}

       & & \multicolumn{5}{c}{TEST Accuracy (\%)} \\
       & & {\bf Day $1$} &  {\bf Day $2$} &  {\bf Day $3$} &  {\bf Day $4$} &  {\bf Average}  \tstrut \bstrut \\
        
       \specialrule{.1em}{.05em}{.0em} 
        
\multirow{5}{*}{\rotatebox{90}{TRAIN \ \ \ }}                     
                    & {\bf Day $1$}     & $67.7$ & $41.9$ & $37.2$ & $67.2$ & $53.5$ \tstrut \bstrut \\
                    & {\bf Day $2$}     & $40.1$ & $67.8$ & $35.4$ & $66.8$ & $57.5$ \tstrut \bstrut \\
                    & {\bf Day $3$}     & $62.0$ & $63.5$ & $66.4$ & $64.9$ & $64.2$ \tstrut \bstrut \\
                    & {\bf Day $4$}     & $62.9$ & $64.1$ & $65.3$ & $67.1$ & $64.8$ \tstrut \bstrut \\
                    & {\bf All Days}  & $73.4$ & $71.0$ & $68.1$ & $68.9$ & $70.3$ \tstrut \bstrut \\
                                         
\end{tabular}
\caption{\emph{\emph{Accuracy of predictors trained on single days compared with a predictor trained on all days together. For a fair comparison, the training dataset have same size ($100$ examples per class).}}}
\label{tab:redundancy}
\end{center}
\end{table}

\subsection{Supervision and clutter}\label{sec:background}

\begin{figure}[t]
\begin{center}
  \includegraphics[width=\columnwidth]{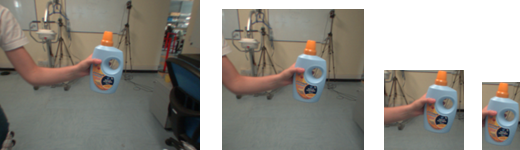}
\end{center}
   \caption{Different supervision strategies evaluated in this work. From left to right: Whole image (no segmentation), large ($220\times220$px) and small ($120\times120$px) bounding boxes cropped around the object of interest using motion detection (see Sec.~\ref{sec:setup}) and manual segmentation.}
\label{fig:crops}
\end{figure}

An important component of the visual recognition framework considered in this work, is the motion detection routine that performs the preliminary crop around the object in the image (see Sec.~\ref{sec:setup}). In this section we investigate the actual impact of such a strategy by comparing it with two other approaches. On one hand, we consider the setting where we take the whole image in input and no crop is performed, in order to understand the benefits of our method; on the other hand, we manually fix the bounding box around the object, to evaluate what could still be gained in terms of performance. Fig.~\ref{fig:crops} reports an example of these strategies, considering the dataset acquired on Day $1$.

In Table~\ref{tab:crops}, we report the classification accuracy of recognition systems trained on images cropped accordingly to the strategies introduced above. We can notice that motion detection provides already a remarkable boost in performance with respect to taking the whole image, thus suggesting that the presence of the background has a disrupting effect. This result is actually surprising considering that the typical training data used for large image retrieval tasks such as ImageNet~\cite{ILSVRCarxiv14}, often depict large portions of the background as in our case. We point out that manual cropping provides further benefits to the classification accuracy. Although this strategy is not applicable to real robotics settings, this result encourages to develop finer approaches to object localization that would eventually lead to similar performance.

\begin{table}[t]
\begin{center}

\begin{tabular}{rrcccc}

       & & \multicolumn{4}{c}{TEST Accuracy (\%)} \\
       & & {\bf Image} & {\bf Crop 1} &  {\bf Crop 2} &  {\bf Manual} \tstrut \bstrut \\
        
       \specialrule{.1em}{.05em}{.0em} 
        
\multirow{4}{*}{\rotatebox{90}{TRAIN \ \ }} 
                      & {\bf Image}                 & $50.6$ & $48.8$  & $36.3$ &  $20.6$ \tstrut \bstrut \\
                       & {\bf Crop 1}           & $50.3$ & $62.2$  & $57.7$ &  $24.9$ \tstrut \bstrut \\
                       & {\bf Crop 2}           & $30.1$ & $50.8$  & $73.9$ &  $28.7$ \tstrut \bstrut \\
                       & {\bf Manual}               & $6.8$   & $8.9$    & $12.2$ &  $81.7$ \tstrut \bstrut \\
                                         
\end{tabular}
\caption{\emph{\emph{Comparison of the classification accuracy achieved by recognition systems trained on iCubWorl28 for different levels of supervision: whole image, crop $1$ ($220\times220$px), crop $2$ ($120\times120$px) and manual segmentation. See Fig.~\ref{fig:crops} for examples.}}}
\label{tab:crops}
\end{center}
\end{table}

\subsection{How many object can iCub recognize?}\label{sec:final_answer}

We finally come back to the original question regarding the maximum number of objects that iCub can recognize with the visual recognition system described in this paper. Specifically, we are interested in achieving human-level performance on {\sc iCubWorld28}, which we set, for reference, to a high value of accuracy: $0.98$. Table~\ref{tab:howmany} provides the answer to this question, returning the maximum number of objects that can be recognized with accuracy of $0.98$ for varying levels of confidence. Overall we observe that only few objects are actually recognized with high confidence. This result shows that modern visual recognition approaches have actually opened the way to answer the ambitious question we asked in this work, but also that the problem is far from being solved.

\begin{table}[t]
\begin{center}

\begin{tabular}{rccccc}

        & \multicolumn{5}{c}{{\bf Confidence}} \\
        & $98\%$ & $90\%$ &$80\%$ & $70\%$ & $50\%$ \tstrut \bstrut \\    

       \specialrule{.1em}{.05em}{.0em} 

       {\bf \# Objects} & {\bf $2$} &  {\bf $4$}  &  {\bf $6$} & {\bf $7$} & {\bf $14$} \tstrut \bstrut \\

\end{tabular}
\caption{\emph{\emph{The maximum number of objects that iCub is able to recognize with $0.98$ accuracy.}}}
\label{tab:howmany}
\end{center}
\end{table}

\subsection{Comparison with other Visual Architectures}\label{sec:comparison_previous_work}

For completeness, we close this work by providing a brief comparison with other methods for visual recognition. In Table~\ref{tab:previous_work} we report the classification accuracy of systems trained/tested on the different days of {\sc iCubWorld28}. The following architectures for visual representation learning were evaluated: Bag of Words (BOW)~\cite{csurka04}, Sparse Coding~\cite{yang09}, Fisher Vector~\cite{Perronnin10}, VLAD~\cite{jegou10}, PHOW~\cite{bosch2007image} and the Overfeat implementation~\cite{sermanet14} of a Convolutional Neural Network. Due to space limitation we refer the reader to the original papers for more informations about these methods. However, we point out that these approaches can be divided in two groups: pre-trained ``deep'' architectures (the CNNs {\it CaffeNet} and {\it OverFeat}) and single layer ``shallow'' representations (the remaining methods), where the ``dictionary learning'' stage was carried out on a subset of the training set of {\sc iCubWorld28}. As can be noticed pre-trained CNNs clearly outperform the others and this was the main reason for the choice of {\it CaffeNet} for our experiments. 

\begin{table}[t]
\begin{center}

\rowcolors{3}{gray!35}{}
\begin{tabular}{rccccc}

       & \multicolumn{5}{c}{TEST Accuracy (\%)} \\
       & {\bf Day $1$} & {\bf Day $2$} & {\bf Day $3$} & {\bf Day $4$} & {\bf Avg.} \tstrut \bstrut \\
        
       \specialrule{.1em}{.05em}{.0em} 
        
        PHOW~\cite{bosch2007image} & $42.5$ & $39.0$ & $34.6$ & $39.0$ & $44.1$ \tstrut \bstrut \\
        BoW~\cite{csurka04} & $44.9$ & $40.8$ & $35.3$ & $38.8$ & $41.1$ \tstrut \bstrut \\
        Sparse Coding~\cite{yang09} & $29.2$ & $24.1$ & $21.9$ & $23.7$ & $30.6$ \tstrut \bstrut \\
        HMAX~\cite{serre07} & $30.5$ & $27.3$ & $25.4$ & $23.7$ & $32.8$ \tstrut \bstrut \\
        Fisher Vectors~\cite{Perronnin10} & $47.3$ & $44.7$ & $41.5$ & $44.3$ & $48.6$ \tstrut \bstrut \\
        VLAD~\cite{jegou10} &$44.2$ & $40.0$ & $35.0$ & $38.1$ & $44.5$ \tstrut \bstrut \\
        {\it CaffeNet}~\cite{jia2014caffe} & $75.9$ & $70.9$ & $71.9$ & $73.9$ & $80.8$ \tstrut \bstrut \\
        {\it OverFeat}~\cite{sermanet14} & $66.8$ & $57.5$ & $57.7$ & $60.0$ & $68.3$ \tstrut \bstrut \\

\end{tabular}
\caption{\emph{\emph{Comparison of several architectures for visual representation learning applied to the visual classification problem of {\sc iCubWorld28}. Modern Convolutional Neural Networks (the {\it CaffeNet} used in this work and {\it Overfeat}) clearly outperform previous methods.}}}
\label{tab:previous_work}
\end{center}
\end{table}

\section{Discussion and Future Work}
\label{sec:conclusion}

In this paper we tested the current visual recognition capabilities of a humanoid robot, the iCub. Our analysis addressed the generic question ``{\it How many objects can (currently) iCub recognize?}'', which was then formulated more accurately as the problem of determining the maximum number of objects that state-of-the-art visual recognition systems can recognize with (virtually) perfect accuracy. 

We identified a natural human-robot interaction application as a possible testbed for our investigation of the visual recognition problem. In order to foster the reproducibility of our experiments, we collected a novel dataset within this scenario, {\sc iCubWorld28}, comprising images depicting $28$ object classes and acquired over the course of $4$ days. 

We approached the problem by first defining a measure performance that would allow us to operatively quantify our confidence that results observed off-line on {\sc iCubWorld28} would then generalize to the real application. We then identified multiple aspects of the robotics context that could be leveraged to improve the overall recognition capabilities of the otherwise purely-visual system. In particular we empirically observed that exploiting the temporal consistency of subsequent frames in the visual stream or adopting weakly-supervised strategies to reduce the amount of distractors in the image can be extremely beneficial. Following these principles we were able to provide a preliminary answer to the original question. Our results show on one hand that modern visual representation architectures such as CNN are finally able to address visual recognition in robotic settings but on the other hand they point out that the problem is extremely challenging and far from being solved.

{\small
\bibliographystyle{ieee}
\bibliography{biblio}
}

\end{document}